\documentclass[conference]{IEEEtran}
\IEEEoverridecommandlockouts
\usepackage{cite}
\usepackage{amsmath,amssymb,amsfonts}
\usepackage{bm}
\usepackage{algorithm}
\usepackage{algorithmic}
\usepackage{graphicx}
\usepackage{textcomp}
\usepackage{xcolor}
\usepackage{multirow}

\def\BibTeX{{\rm B\kern-.05em{\sc i\kern-.025em b}\kern-.08em
    T\kern-.1667em\lower.7ex\hbox{E}\kern-.125emX}}

\begin{document}

\title{Random Padding Data Augmentation\\

}

\author{\IEEEauthorblockN{Nan Yang, Laicheng Zhong, Fan Huang, Dong Yuan and Wei Bao}
\IEEEauthorblockA{\textit{Faculty of Engineering, The University of Sydney} \\
\{n.yang, laicheng.zhong, fan.huang, dong.yuan, wei.bao\}@sydney.edu.au}
}

\maketitle

\begin{abstract}
The convolutional neural network (CNN) learns the same object in different positions in images, which can improve the recognition accuracy of the model. An implication of this is that CNN may know where the object is. The usefulness of the features' spatial information in CNNs has not been well investigated. In this paper, we found that the model's learning of features' position information hindered the learning of the features' relationship. Therefore, we introduced Random Padding, a new type of padding method for training CNNs that impairs the architecture's capacity to learn position information by adding zero-padding randomly to half of the border of feature maps. Random Padding is parameter-free, simple to construct, and compatible with the majority of CNN-based recognition models. This technique is also complementary to data augmentations such as random cropping, rotation, flipping and erasing, and consistently improves the performance of image classification over strong baselines.
\end{abstract}


\section{Introduction}
Convolutional Neural Network (CNN) is an important component in computer vision and plays a key role in deep learning architecture, which extracts low/mid/high-level features \cite{zeiler2014visualizing} and classifiers naturally in an end-to-end multilayer, and the "levels" of features can be evolved by the depth of stacked layers. The idea of CNN model design is inspired by live organisms' inherent visual perception process \cite{hubel1962receptive}. 
The shallow layers learn the local features of the image, such as the color and geometric shape of the image object, while the deep layers learn more abstract features from the input data, such as contour characteristics and other high-dimensional properties.  The multi-layer structure of the CNN can automatically learn features' spatial information from the input image data. Spatial information is represented by matrices in the hierarchical CNN models, which is the conversion between the overall coordinate system of the object and the coordinate system of each component. 
Thus, the spatial information of the features learned by CNN can perform shift-invariant classification of input information \cite{zhang1996improved}.







In fact, what really provides free translation and deformation invariance for the CNN architecture are the convolution structure and pooling operation, while the use of sub-sampling and stride will break this special characteristic \cite{azulay2019deep}. 
Even if the target position is changed, the operations of convolution and pooling can still extract the same information from images, and then flatten it to the same feature value in different orders of the following fully connected layer. However, sub-sampling and stride reduce part of the information of the input image, which leads to the loss of some features, thus breaking the translation invariance of CNN.
Recent research on data argumentation has attempted to enhance the invariance by performing the operations of translation, rotation, reflection and scaling \cite{bruna2013invariant,sifre2013rotation,cohen2016group,worrall2017harmonic,esteves2018learning} on the input images. However, it cannot really improve the model's learning ability of shift invariance. 

The shift-invariant ability of CNN depends on the learning of features' spatial information, which contains two types of information, i.e., \textit{features relationship} and \textit{position information}. \textit{Features relationship} refers to the relative position among different features, while \textit{position information} represents the absolute position of features in the image. 
We deem that features relationship is helpful in CNN, as if a feature is useful in one image location, the same feature is likely to be useful in other locations. The Capsule Network \cite{kosiorek2019stacked} is designed to learn features relationship from images, i.e., the spatial relationships between whole objects and their parts. However, it is difficult to implement on complex datasets, e.g., CIFAR-10 and ImageNet. On the other hand, we believe position information is harmful to CNN, as learning it will impede the model's acquisition of features relationship. Recent evidence suggests that position information is implicitly encoded in the extracted feature maps, thus 
non-linear readout of position information further augments the readout of absolute position \cite{islam2019much}. Additionally, these studies point to that zero-padding and borders serve as an anchor for spatial information that is generated and ultimately transmitted over the whole image during spatial abstraction. Hence, how to reduce the position information introduced by zero-padding has been long-ignored in CNN solutions to vision issues.

\begin{figure}[t]
\centering
\includegraphics[width=0.9\columnwidth]{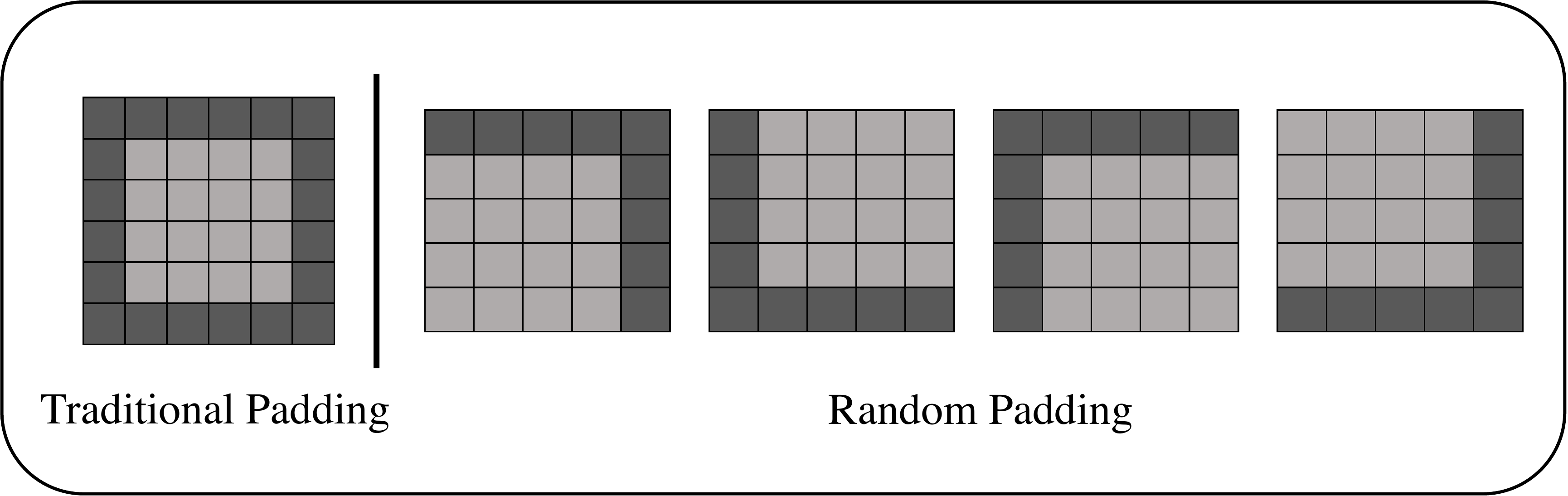} 
\caption{Traditional Padding and Random Padding.}
\label{fig1}
\end{figure}

In this paper, with the purpose of reducing CNN's learning of position information, we proposed the Random Padding operation shown in Figure \ref{fig1}, that is a variant of traditional padding technology. Random Padding is to add zero-padding to the randomly chosen half boundary of the feature maps, which will weaken the position information and let the CNN model better understand the features relationship. This technique makes CNN models more robust to the change of object's absolute position in the images.

The contribution of the paper is summarised as follows:
\begin{itemize}
\item We investigate the usefulness of spatial information in CNN and propose the new approach to improving the model accuracy, i.e., through reducing the position information in CNN.
\item We propose the Random Padding operation that can be directly added to a variety of CNN models without any changes to the model structure. It is lightweight that requires no additional parameter learning or computing in model training. As the operation does not make changes on the input images, it is complementary to the traditional data augmentations used in CNN training. 
\item We conduct extensive experiments on popular CNN models. The results show that the Random Padding operation can reduce the extraction of the position information in CNN models, and improve the accuracy on image classification. Comprehensive experiment results and the complete program code are in supplementary materials.
\end{itemize} 

\section{Related Work}

\subsection{Approaches to Improve Accuracy of CNNs}

The structure evolution of the convolutional neural networks has gradually improved the accuracy, i.e., Alexnet with ReLu and Dropout \cite{krizhevsky2012imagenet}, VGG with ${3 * 3}$ kernels \cite{DBLP:journals/corr/SimonyanZ14a}, Googlenet with inception \cite{szegedy2015going} and Resnet with residual blocks \cite{he2016deep}. In addition to the upgrade of CNN architectures, the augmentation of input data is also an indispensable part of improving performance.

\noindent\textbf{Data Augmentation.} The common demonstrations showing the effectiveness of data augmentation come from simple transformations, such as translation, rotation, flipping, cropping, adding noises, etc, which aim at artificially enlarging the training dataset \cite{shorten2019survey}. The shift invariance of the object is encoded by CNNs to improve the model's learning ability for image recognition tasks. For example, rotation augmentation on source images is performed by randomly rotating clockwise or counterclockwise between 0 and 360 degrees with the center of the image as the origin, reversing the entire rows and columns of image pixels horizontally or vertically is called flipping augmentation, and random cropping is a method to reduce the size of the input and create random subsets of the original input \cite{krizhevsky2012imagenet}. Random Erasing \cite{zhong2020random} is another interesting data augmentation technique that produces training images with varying degrees of occlusion.

\noindent\textbf{Other Approaches.}
In addition to geometric transformation and Random Erasing, there are many other image manipulations, such as noise injection, kenel filters, color space transformations and mixing images \cite{shorten2019survey}. Noise injection is injecting a matrix of random values usually drawn from a Gaussian distribution, which can help CNNs learn
more robust features \cite{moreno2018forward}. Kernel filters are a widely used image processing method for sharpening and blurring images \cite{kang2017patchshuffle}, whereas color space transformations aims to alter the color values or distribution of images \cite{DBLP:journals/corr/ChatfieldSVZ14,jurio2010comparison}. Mixing images appeared in recent years, which has two approaches. The one is cropping images randomly and concatenate the croppings together to form new images \cite{inoue2018data}, the other is using non-linear methods to combine images to create new training examples \cite{summers2019improved}.

Different from the above approaches, we design the Random Padding operation to improve the performance of CNN models from the perspective of reducing the position information in the network.
\subsection{Padding in CNN}
The boundary effect is a well-researched phenomenon in biological neural networks \cite{tsotsos1995modeling,sirovich1979effect}. Previous research has addressed the boundary effect for artificial CNNs by using specific convolution filters for the border regions \cite{innamorati2019learning}. At some point during the convolution process, the filter kernel will come into contact with the image border \cite{islam2019much}. Classic CNNs use zero-padding to enlarge the image for filtering by kernels. The cropped images are filled by paddings to reach the specified size \cite{xie2018mitigating}. Guilin Liu and his colleagues proposed a simple and effective padding scheme, called partial convolution-based padding. Convolution results are reweighted near image edges relying on the ratios between the padded region and the convolution sliding window area \cite{liu2018partial}.

Padding is an additional pixels that can be added to the border of an image. In the process of convolution, the pixel in the corner of an image will only get covered one time by kernels but the middle pixel will get covered more than once basically, which will cause shrinking outputs and losing information on corners of the image. Padding works by extending the area in which a convolution neural network processes an image. Padding is added to the frame of the image to enlarge the image size for the kernel to cover better, which assists the kernel with processing the image. Adding padding operations to a CNN help the model get a more accurate analysis of images.
\section{Random Padding for CNN}
This section presents the Random Padding operation for training in the convolutional neural network (CNN). Firstly, we introduce the detailed procedure of Random Padding. Next, using comparative experiments to verify that the extraction of position information will be reduced in CNNs with the method of Random Padding. Finally, the implementation of Random Padding in different CNN models is introduced. 

\subsection{Random Padding Operation}\label{AA}
In CNN training, we replace the traditional  padding with the technique of Random Padding, which has four types of padding selections shown in Figure \ref{fig1}. For the feature maps generated in the network, Random Padding will perform the zero-padding operation randomly on the four boundaries according to the required thickness of the padding by the feature maps.
When padding thickness equals 1, Random Padding will first randomly select one padding of the four patterns. Assuming that the size of the feature map in the training is ${w * w}$, the feature map will become ${(w+1)*(w+1)}$ after this new padding method. Then, Random Padding will randomly select one of the four modes again, which will change the size of the feature map to $((w+1)+1)*((w+1)+1)$. In general, Random Padding will perform $2n$ padding selections if the padding thickness is $n$, where $n=1,2,3$ are most common in the CNN models. The detailed steps of Random Padding are shown in Algorithm \ref{alg:1}. In this process, the position of features will be randomly changed by adding Random Padding, hence the learning of the object's position information by CNN will be reduced.
\begin{algorithm}[h]
  \caption{Random Padding Procedure} 
  \label{alg:1}
  \begin{algorithmic}[1]
    \STATE {\bfseries Input:}
      Input feature map: $I$; The thickness of padding: $n$; The random padding thickness of four boundaries, left, right, top and bottom: $l, r, t, b$; Padding options: $S$
    \STATE {\bfseries Output:}
       Feature map with padding $I^*$
       \STATE$l,\,r,\,t,\,b \gets 0$;
       \STATE$S \gets[[1,0,1,0],[1,0,0,1],[0,1,1,0],[0,1,0,1]]$;
       \FOR{$i\leftarrow 1$,\, $2n$}
       \STATE $P_r[] \gets (S,\,1)$
       \COMMENT{select a padding option randomly}
       \STATE $l \gets l + P_r[0]$
       \COMMENT{padding\_left}
       \STATE $r \gets r + P_r[1]$
       \COMMENT{padding\_right}
       \STATE $t \gets t + P_r[2]$
       \COMMENT{padding\_top}
       \STATE $b \gets b + P_r[3]$
       \COMMENT{padding\_bottom}
       \ENDFOR
       \STATE $I^* \gets ([l,\,r,\,t,\,b])(I)$\

  \end{algorithmic}
\end{algorithm}

\subsection{Validation Method for Position Information Reduction in CNNs}

The position information has been proved to be implicitly encoded in the feature map extracted by CNN, which was introduced by the traditional zero padding \cite{islam2019much}. In this article, we proposed the hypothesis that the Random Padding operation will reduce the extraction of position information in CNNs. In this sub-section, we prove this hypothesis by comparing position information in an end-to-end manner between the CNN with traditional padding and the CNN with Random Padding.

\begin{figure}[t]
\centering
\includegraphics[width=1.0\columnwidth]{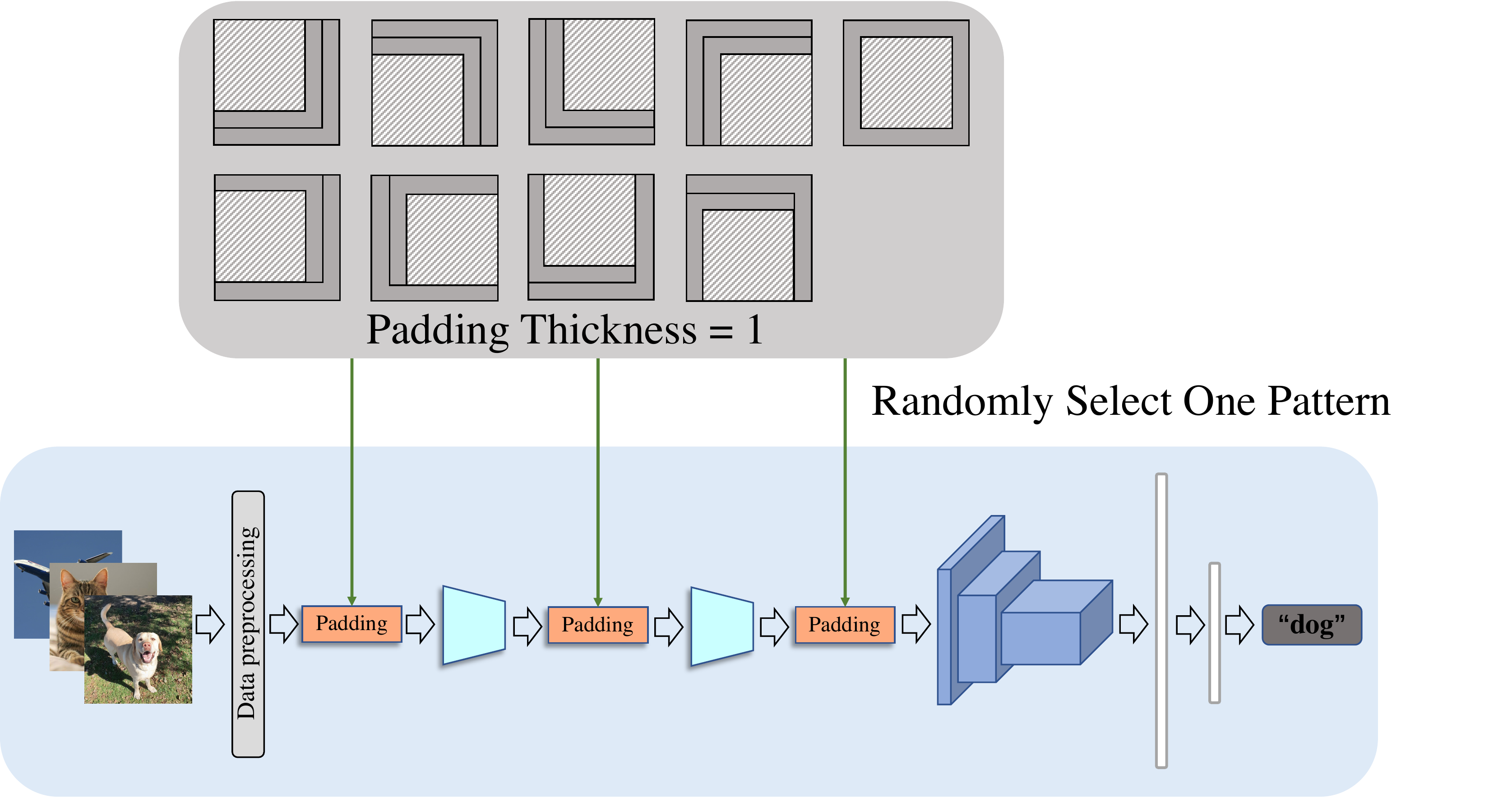} 
\caption{Random Padding in the CNN.}
\label{fig2}
\end{figure}

\begin{figure*}[t]
\centering
\includegraphics[width=0.95\textwidth]{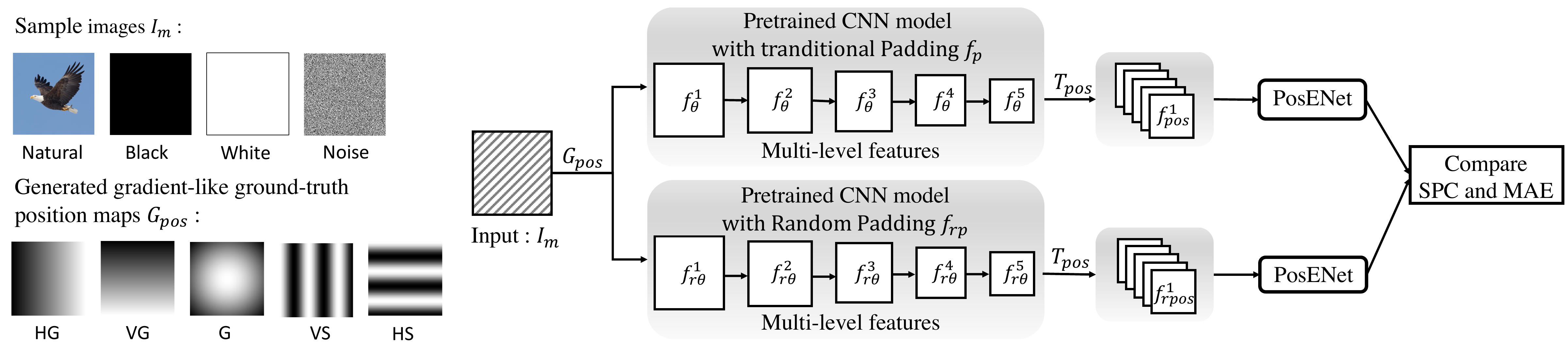} 
\caption{Compare position information between the CNN with traditional padding and the CNN with Random Padding.}
\label{fig3}
\end{figure*}

This validation experiment used the Position Encoding Network \cite{islam2019much}, which is composed of two critical components: a feedforward convolutional encoder network and a simple position encoding module. For this task, the two encoder networks collected characteristics, from the CNN with traditional padding and the CNN with Random Padding, at different layers of abstraction. After the collection of multi-scale features from the front networks as inputs, the other two position encoding modules outputted the prediction of position information. Due to the fact that position encoding measurement is a novel concept, there is no universal metric. We used Spearman's rank correlation coefficient (SPC) and Mean absolute error (MAE) proposed by \cite{islam2019much} to evaluate the position encoding performance to verify that the Random Padding operation reduces the amount of position information extracted by CNNs. The higher the SPC, the higher the correlation between the output and the ground truth map, while the MAE is the opposite. We will present the detailed experiment setting and results in Section 4.

\subsection{Construct CNN with Random Padding}
The Random Padding operation can be added to different types of backbone networks to construct CNN for image classification. 
In order to better analyze the relationship between the way of padding joins CNN and the improvement of model performance,
we replaced the traditional padding of the first one, the first two and the first three padding layers with the Random Padding operation to compare the accuracy of image classification in various CNN models, which is shown in Figure \ref{fig2}. Since the Random Padding operation is complementary to general data augmentation methods, we employ random cropping, random flipping, random rotation, and random Erasing methods to enrich the training datasets. We will present the detailed experiment setting and results in Section 5. 
\section{Evaluation of Position Information in CNNs}

This section quantitatively evaluates the impact of CNN using traditional padding and Random padding on position information extraction. The first part introduces the dataset used by prertrained models and the evaluation metrics for position information. Secondly, we compare the learning ability of CNN with traditional padding and CNN with Random Padding on position information extraction. In the third part, we analyze the experimental results and justify that the CNN with Random Padding reduces the extraction of position information.

\subsection{Dataset and Evaluation Metrics}
\noindent\textbf{Dataset}
We use the Imagenet dataset \cite{he2019rethinking} to train the basic VGG and VGG with Random Padding as our initialization networks, and then use the DUT-S dataset \cite{wang2017learning} as our training set, which contains 10,533 images for training. Following the common training protocol used in \cite{zhang2017amulet,liu2018picanet}, we train the models on the same training set of DUT-S and evaluate the existence of position information on the synthetic images (white, black, and Gaussian noise) and the natural image from the website of News-Leader. Notably, we adhere to the standard setting used in saliency detection to ensure that the training and test sets do not overlap. Since the position information is largely content-independent, any image or model can be used in our studies.

\noindent\textbf{Evaluation}
At present, there is no universal standard for the measurement of position encoding, so we evaluate the performance of position information according to the two different methods for metrics (Spearman's rank correlation coefficient (SPC) and Mean Absolute Error (MAE)) previously used by \cite{islam2019much}. SPC is a non-parametric measurement of the association between the ground-truth and the predicted position map. We maintain the SPC score within the range [-1, 1] to facilitate understanding. MAE measures the average magnitude of the errors in the predicted position map and the ground-truth gradient position map, without considering their direction. 
The lower the SPC value, the less position information the model produces, and the higher the MAE value, the less position information the model outputs.
We expect lower SPC values and higher MAE values after applying the Random Padding operation in CNNs.
\begin{table*}[htbp]
\caption{Comparison of SPC and MAE in CNNs using traditional padding or Random Padding across different image types}
\centering
\small


\begin{tabular}{|l|l|ll|ll|ll|ll|}
\hline
                    &         & \multicolumn{2}{c|}{Natrue}                            & \multicolumn{2}{c|}{Black}                             & \multicolumn{2}{c|}{White}                             & \multicolumn{2}{c|}{Noise}                             \\ \hline
                    & Model   & \multicolumn{1}{c|}{SPC}    & \multicolumn{1}{c|}{MAE} & \multicolumn{1}{c|}{SPC}    & \multicolumn{1}{c|}{MAE} & \multicolumn{1}{c|}{SPC}    & \multicolumn{1}{c|}{MAE} & \multicolumn{1}{c|}{SPC}    & \multicolumn{1}{c|}{MAE} \\ \hline
\multirow{3}{*}{H}  & PosENet & \multicolumn{1}{l|}{0.130}  & 0.284                    & \multicolumn{1}{l|}{0.}     & 0.251                    & \multicolumn{1}{l|}{0.}     & 0.251                    & \multicolumn{1}{l|}{0.015}  & 0.251                    \\ \cline{2-10} 
                    & VGG     & \multicolumn{1}{l|}{0.411}  & 0.239                    & \multicolumn{1}{l|}{0.216}  & 0.242                    & \multicolumn{1}{l|}{0.243}  & 0.242                    & \multicolumn{1}{l|}{0.129}  & 0.245                    \\ \cline{2-10} 
                    & VGG\_RP & \multicolumn{1}{l|}{-0.116} & 0.253                    & \multicolumn{1}{l|}{0.021}  & 0.252                    & \multicolumn{1}{l|}{0.023}  & 0.255                    & \multicolumn{1}{l|}{-0.045} & 0.251                    \\ \hline
\multirow{3}{*}{V}  & PosENet & \multicolumn{1}{l|}{0.063}  & 0.247                    & \multicolumn{1}{l|}{0.063}  & 0.254                    & \multicolumn{1}{l|}{0.}     & 0.253                    & \multicolumn{1}{l|}{-0.052} & 0.251                    \\ \cline{2-10} 
                    & VGG     & \multicolumn{1}{l|}{0.502}  & 0.234                    & \multicolumn{1}{l|}{0.334}  & 0.242                    & \multicolumn{1}{l|}{0.433}  & 0.247                    & \multicolumn{1}{l|}{0.120}  & 0.250                    \\ \cline{2-10} 
                    & VGG\_RP & \multicolumn{1}{l|}{-0.174} & 0.249                    & \multicolumn{1}{l|}{-0.174} & 0.249                    & \multicolumn{1}{l|}{-0.027} & 0.257                    & \multicolumn{1}{l|}{0.100}  & 0.249                    \\ \hline
\multirow{3}{*}{G}  & PosENet & \multicolumn{1}{l|}{0.428}  & 0.189                    & \multicolumn{1}{l|}{0.}     & 0.196                    & \multicolumn{1}{l|}{0.}     & 0.206                    & \multicolumn{1}{l|}{0.026}  & 0.198                    \\ \cline{2-10} 
                    & VGG     & \multicolumn{1}{l|}{0.765}  & 0.14                     & \multicolumn{1}{l|}{0.421}  & 0.205                    & \multicolumn{1}{l|}{0.399}  & 0.192                    & \multicolumn{1}{l|}{0.161}  & 0.187                    \\ \cline{2-10} 
                    & VGG\_RP & \multicolumn{1}{l|}{-0.49}  & 0.200                    & \multicolumn{1}{l|}{-0.009} & 0.196                    & \multicolumn{1}{l|}{-0.040} & 0.195                    & \multicolumn{1}{l|}{-0.051} & 0.196                    \\ \hline
\multirow{3}{*}{HS} & PosENet & \multicolumn{1}{l|}{0.187}  & 0.306                    & \multicolumn{1}{l|}{0.}     & 0.308                    & \multicolumn{1}{l|}{0.}     & 0.308                    & \multicolumn{1}{l|}{-0.060} & 0.308                    \\ \cline{2-10} 
                    & VGG     & \multicolumn{1}{l|}{0.234}  & 0.211                    & \multicolumn{1}{l|}{0.227}  & 0.297                    & \multicolumn{1}{l|}{0.285}  & 0.301                    & \multicolumn{1}{l|}{0.253}  & 0.292                    \\ \cline{2-10} 
                    & VGG\_RP & \multicolumn{1}{l|}{0.043}  & 0.308                    & \multicolumn{1}{l|}{0.049}  & 0.310                    & \multicolumn{1}{l|}{-0.127} & 0.306                    & \multicolumn{1}{l|}{-0.066} & 0.309                    \\ \hline
\multirow{3}{*}{VS} & PosENet & \multicolumn{1}{l|}{0.015}  & 0.315                    & \multicolumn{1}{l|}{0.}     & 0.308                    & \multicolumn{1}{l|}{0.}     & 0.313                    & \multicolumn{1}{l|}{-0.022} & 0.310                    \\ \cline{2-10} 
                    & VGG     & \multicolumn{1}{l|}{0.339}  & 0.292                    & \multicolumn{1}{l|}{0.240}  & 0.296                    & \multicolumn{1}{l|}{0.229}  & 0.299                    & \multicolumn{1}{l|}{0.249}  & 0.292                    \\ \cline{2-10} 
                    & VGG\_RP & \multicolumn{1}{l|}{0.004}  & 0.308                    & \multicolumn{1}{l|}{0.040}  & 0.308                    & \multicolumn{1}{l|}{0.026}  & 0.308                    & \multicolumn{1}{l|}{0.050}  & 0.308                    \\ \hline
\end{tabular}
\label{table1}
\end{table*}

\subsection{Architectures and Settings}
\noindent\textbf{Architectures}
We first build two pre-trained networks based on the basic architecture of VGG with 16 layers. The first network uses traditional padding, and the second one applies the technique of Random Padding on the first two padding layers. The proper number of padding layers added to the CNN is analyzed in Section 5. Meanwhile, we construct a randomization test by using a normalized gradient-like position map as the ground-truth. The generated gradient-like ground-truth position maps contain Horizontal gradient (HG) and vertical gradient (VG) masks, horizontal and vertical stripes (HS, VS), and Gaussian distribution (G). As shown in Figure \ref{fig3}, the combination of natural images $I_m\in\mathbb{R}^{h\times w\times 3} $ and gradient-like masks $G_{pos}\in\mathbb{R}^{h\times w}$ is used as the input of two pretrained models with fixed weights. We remove the average pooling layer and the layer that assigns categories of the pretrained model to construct an encoder network $f_p$ and $f_{rp}$ for extracting feature maps. The features ($f^1_\theta,f^2_\theta,f^3_\theta,f^4_\theta,f^5_\theta$) and ($f^1_{r\theta},f^2_{r\theta},f^3_{r\theta},f^4_{r\theta},f^5_{r\theta}$) we extract from the two encoder networks respectively come from five different abstraction layers, from shallow to deep. The following is a summary of the major operations:

\begin{equation}
    \begin{aligned}
        f^i_\theta=\bm{W}_p \ast I_m(G_{pos})\\
        f^i_{r\theta}=\bm{W}_{rp} \ast I_m(G_{pos})
    \end{aligned}
\end{equation}
where $\bm{W}_p$ denotes frozen weights from the model using traditional padding, and $\bm{W}_{rp}$ represents frozen weights from the model using the Random Padding operation. $\ast$ indicates the model operation.

After multi-scale features collection, the transformation function $T_{pos}$ performs bi-linear interpolation on the extracted feature maps of different sizes to create feature maps with the same spatial dimension. ($f^1_{pos},f^2_{pos},f^3_{pos},f^4_{pos},f^5_{pos}$) and ($f^1_{rpos},f^2_{rpos},f^3_{rpos},f^4_{rpos},f^5_{rpos}$) can be summarized as:

\begin{align}
\begin{aligned}
f^i_{pos}=T_{pos}(f^i_\theta)\\
f^i_{rpos}=T_{pos}(f^i_{r\theta})
\end{aligned}
\end{align}

These resized feature maps should be concatenated together and then send into Position Encoding Module (PosENet) \cite{islam2019much}, which only has one convolutional layer. The features are delivered to PosENet and trained, where the goal of this training is to generate a pattern that is only related to the position information and has nothing to do with other features. It should be noted that during the training process, the parameters of pretrained networks are fixed. The final stage of this study is to compare the extraction of the amount of position information between the CNN with traditional padding and the CNN with Random Padding.

\noindent\textbf{Settings}
The models we choose to compare the position information of feature maps are traditional VGG16 and VGG16 with Random Padding. We initialize the CNN models by pre-training on the ImageNet dataset and keep the weights frozen in our comparison experiment. The size of the input image should be 224×224, which can be a natural picture, a black, a white, or a noise image.  We also apply five different ground-truth patterns HG, VG, G, HS and VS, which represent horizontal and vertical gradients, 2D Gaussian distribution, horizontal and vertical stripes respectively. All feature maps of different five layers extracted from pre-trained models are resized to a size of 28x28. After the feature map is used as input, the Position Encoding Module (PosENet) will be trained with stochastic gradient descent for 15 epochs with a momentum factor of 0.9, and weight decay of $10^{-4}$. For this task, PosENet only has one convolutional layer with a kernel size of 3 × 3 without any padding, which will learn position information directly from the input.

\begin{figure}[t]
\centering
\includegraphics[width=0.8\columnwidth]{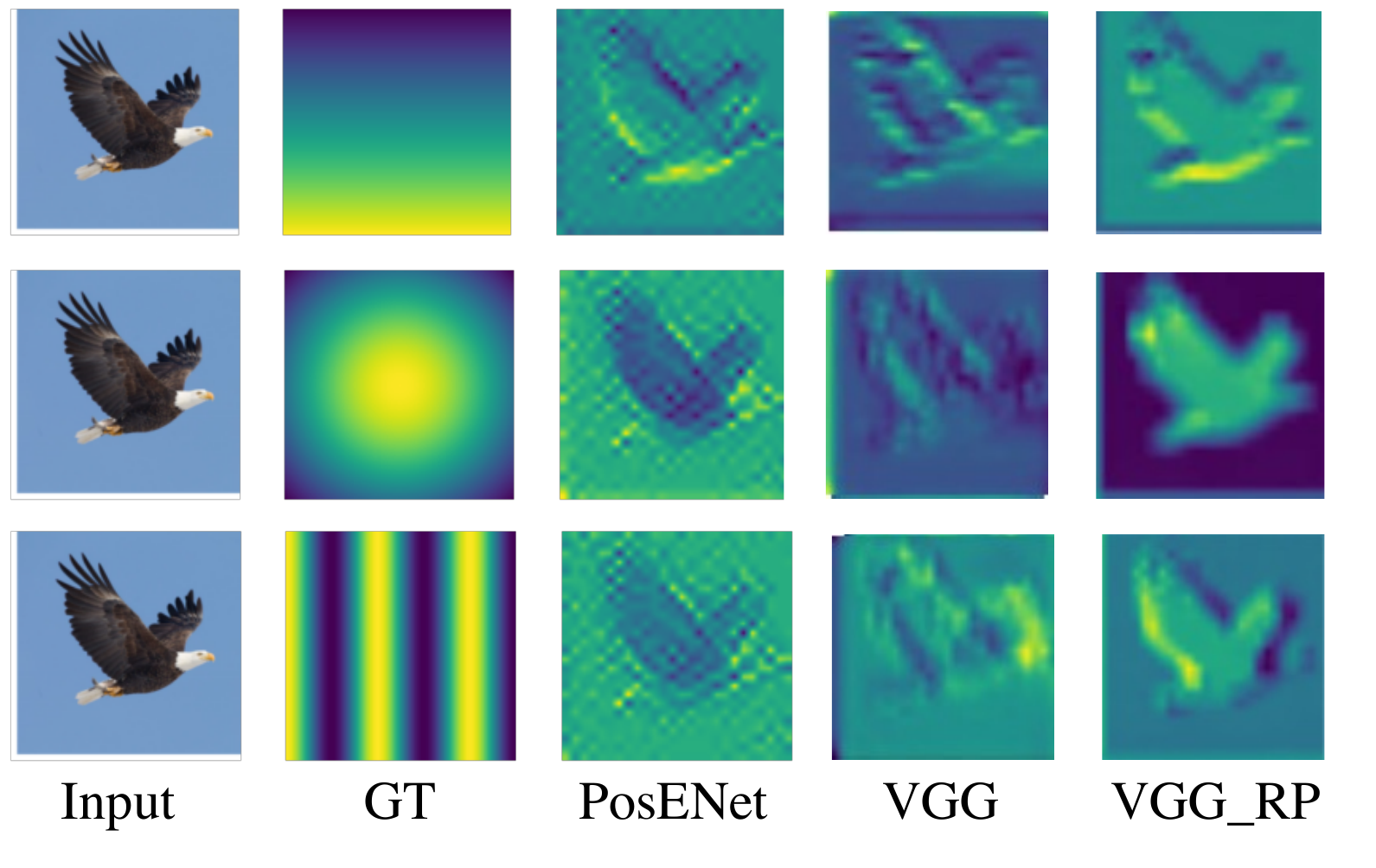} 
\caption{Results of PosENet based networks with traditional padding or Random Padding corresponding to different ground-truth (GT) patterns.}
\label{fig4}
\end{figure}

\begin{table*}[htbp]
\caption{Test errors (\%) with different architectures on CIFAR-10, CIFAR-100 and Fashion-MNIST. \textbf{Baseline:} Baseline model, \textbf{RP\_1:} Random Padding on the first padding layer, \textbf{RP\_2:} Random Padding on the first two padding layers, \textbf{RP\_3:} Random Padding on the first three padding layers.}
\centering
\small
\scalebox{1.0}{
\renewcommand\tabcolsep{3.0pt}}

\begin{tabular}{|l|l|l|l|l|c|}
\hline
Model                          &          & \multicolumn{1}{c|}{Alexnet} & \multicolumn{1}{c|}{VGG16} & \multicolumn{1}{c|}{Googlenet} & Resnet18                          \\ \hline
\multirow{4}{*}{CIFAR10}       & Baseline & 15.19 ± 0.07                 & 12.41 ± 0.05               & 11.47 ± 0.08                   & \multicolumn{1}{l|}{12.08 ± 0.04} \\ \cline{2-6} 
                               & RP\_1    & 13.39 ± 0.05                 & 11.34 ± 0.07               & 10.37 ± 0.09                   & \multicolumn{1}{l|}{8.21 ± 0.07}  \\ \cline{2-6} 
                               & RP\_2    & 12.93 ± 0.12                 & 10.54 ± 0.04               & 10.25 ± 0.07                   & -                                 \\ \cline{2-6} 
                               & RP\_3    & 12.75 ± 0.07                 & 10.61 ± 0.05               & 10.20 ± 0.13                   & -                                 \\ \hline
\multirow{4}{*}{CIFAR100}      & Baseline & 44.57 ± 0.06                 & 46.93 ± 0.04               & 34.81 ± 0.07                   & \multicolumn{1}{l|}{36.90 ± 0.09} \\ \cline{2-6} 
                               & RP\_1    & 42.36 ± 0.07                 & 42.32 ± 0.06               & 34.50 ± 0.07                   & \multicolumn{1}{l|}{31.29 ± 0.08} \\ \cline{2-6} 
                               & RP\_2    & 41.54 ± 0.06                 & 40.66 ± 0.05               & 33.12 ± 0.09                   & -                                 \\ \cline{2-6} 
                               & RP\_3    & 40.66 ± 0.05                 & 39.66 ± 0.04               & 32.94 ± 0.06                   & -                                 \\ \hline
\multirow{4}{*}{Fashion-MNIST} & Baseline & 12.74 ± 0.06                 & 7.02 ± 0.08                & 6.48 ± 0.09                    & \multicolumn{1}{l|}{5.80 ± 0.04}  \\ \cline{2-6} 
                               & RP\_1    & 12.74 ± 0.06                 & 5.49 ± 0.05                & 5.88 ± 0.07                    & \multicolumn{1}{l|}{5.64 ± 0.07}  \\ \cline{2-6} 
                               & RP\_2    & 9.59 ± 0.05                  & 5.49 ± 0.06                & 5.82 ± 0.05                    & -                                 \\ \cline{2-6} 
                               & RP\_3    & 9.40 ± 0.06                  & 5.72 ± 0.08                & 5.94 ± 0.07                    & -                                 \\ \hline
\end{tabular}

\label{table2}
\end{table*}

\subsection{Comparison and Evaluation}
We first use Random Padding on the first two padding layers in one model and then conduct experiments to verify and compare the differences in the position information encoded in the two pretrained models. Following the same protocol, we train networks, based on traditional VGG16 and VGG16 with Random Padding, on each type of ground-truth, and report the experimental results in Table \ref{table1}. In addition, we present the result as a reference by training PosENet without using any pretrained models' feature maps as input. For this task, not only the original image was used, but also pure black, pure white, and Gaussian noise images were used as inputs of PosENet. This is to verify whether the feature contains position information when there is no semantic information. The structure of PosENet is very simple, which can only read out the input features. If the input feature contains more position information, the output image can better approximate the target object; if the input feature does not contain any position information, the output feature map is similar to random noise and cannot output a regular pattern, which represents that the position information is not derived from prior knowledge of the object. Our experimental results are not to evaluate the performance of the model, but to compare the impact of different ways of padding on the position information encoded by the CNN model.

Our experiment takes three kinds of features as input, which are the feature maps extracted by VGG16 with traditional padding, VGG16 with Random Padding, and the natural image without any processing, which are recorded as VGG, VGG\_RP, and PosENet respectively in the Table \ref{table1}. The reason why the original image is used as input is that this kind of image does not contain position information, which shows PosENet's own ability to extract position information and plays a comparative role to other results. PosENet can easily extract the position information from the pretrained VGG models while it is difficult to extract position information directly from the original image. Only when combined with a deep neural network can this network extract position information that is coupled with the ground-truth position map. Prior studies that have noted the traditional zero-padding delivers position information for convolutional neural networks to learn.

According to the results in Table \ref{table1}, VGG\_RP is lower than VGG on the evaluation index SPC value, and almost all higher than VGG on MAE value. Sometimes, VGG\_RP is even lower than PosENet's SPC value. The qualitative results for CNNs with traditional padding or Random Padding across different patterns are shown in Figure \ref{fig4}. The first two columns are the patterns of the input image and the target; the third column is the visualization of directly inputting the source image into PosENet; the fourth and fifth columns are the generation effects of VGG and VGG\_RP respectively. We can observe a connection between the predicted and ground-truth position maps for the VG, G, and VS patterns, indicating that CNNs with Random Padding can better learn the object itself that needs to be recognized, which means the technology of Random Padding can indeed reduce CNN's extraction of position information effectively. Therefore, the basic CNN network will learn the position information of the object, while both the PosENet with only one convolution layer and the CNN with Random Padding hardly learn the position information from the input image.
\section{Evaluation of Random Padding}
This section evaluates the Random Padding operation in CNN for improving the accuracy of image classification.
\subsection{Dataset and Evaluation Metrics}
\noindent\textbf{Dataset} We use three image datasets to train the CNN model, including the well-known datasets CIFAR-10 and CIFAR-100, and a grayscale clothing dataset Fashion-Mnist. There are 10 classes in the dataset of CIFAR-10, and each has 6000 32x32 color images. The training set has 50000 images and the test set has 10000 images. The CIFAR-100 is just like the CIFAR-10, except it has 100 classes containing 600 images each. Fashion-MNIST is a dataset that contains 28x28 grayscale images of 70,000 fashion items divided into ten categories, each with 7,000 images. The training set has 60,000 pictures, whereas the test set contains 10,000. The image size and data format of Fashion-MNIST are identical to those of the original MNIST.

\noindent\textbf{Evaluation Metrics}
The test error assessment is an important part of any classification project, which compares the predicted result of the classified image with its ground truth label. For image classification, test error is used to calculate the ratio of incorrectly recognized images to the total number of images that need to be recognized.
 
\subsection{Experiment Setting}
We use CIFAR-10, CIFAR-100 and Fashion-Mnist to train four CNN architectures, which are Alexnet, VGG, Googlenet and Resnet. We use 16-layer network for VGG and 18-layer network for Resnet. The models with different layers' Random Padding were training for 200 epochs with the learning rate of $10^{-3}$. In our first experiment, we compare the different CNN models trained with Random Padding on different layers. For the same deep learning architecture, all the models are trained from the same weight initialization and all the input data has not been augmented. The second experiment is to apply the Random Padding operation and various data augmentations (e.g., flipping, rotation, cropping and erasing) in the CNNs together to justify the complementary performance of the Random Padding operation with data augmentation methods.
\subsection{Classification accuracy on different CNNs}
The experiments in Section 4 proved the Random Padding operation can reduce the extraction of position information, but did not show how much this method can improve the model performance. So we design a comparative experiment shown in Table \ref{table2} to illustrate the results of the Random Padding operation on different padding layers of different CNN models training on the datasets of CIFAR-10, CIFAR-100 and Fashion-MNIST. For each kind of CNN architecture, we apply the Random Padding operation on the first padding layer,  the first two padding layers and the first three padding layers. Specially, we only replace the traditional padding with Random Padding on the first padding layer of the Resnet18 due to the unique structure of shortcut in Resnet. Based on the principle of controlling variables, we train and test the basic CNN architectures on the same dataset. All the results are shown in Table \ref{table2}.

For CIFAR-10, our approach obtains a 12.75\% error rate using Alextnet with Random Padding on the first three padding layers. The error rate on VGG16 with two padding layers of Random Padding achieves 10.54\%, which improves the accuracy by 1.87\%. Replacing the first three traditional paddings with Random Padding makes Googlenet reach a new state-of-the-art test error of 10.20\%. This method also works well on the architecture of Resnet18, which is 3.87\% higher than the baseline.

For CIFAR-100, all types of models using Random Padding have improved their recognition rates. The Alexnet with three Random Padding layers achieves 40.66\%, which increases the accuracy by 3.91\%. There is an upward trend of accuracy when adding the Random Padding operation in more padding layers in the VGG16 and Googlenet. The test errors on VGG16 and Googlenet are 39.66\% and 32.94\% respectively. The Resnet18 with the technique of Random Padding has improved the accuracy significantly, reaching 68.71\%.

For Fashion-Mnist, although the recognition rate of all models is already very high, the addition of the Random Padding operation still improves the performance of the model.  Alexnet, which has a small number of layers, uses Random Padding to increase its recognition rate by 3.34\% on this simple dataset. VGG16 and Googlenet using two layers of the Random Padding operation have better results than using three layers of the Random Padding operation, reaching 5.49\% and 5.82\%, respectively. The error rate of Resnet with a layer of the Random Padding operation is only reduced by 0.16\%.

In general, as the number of layers deepens, the extracted object features gradually become abstracted, and the encoding of position information also becomes more implicit. The addition of the Random Padding operation in subsequent deep layers will jeopardize the models' learning of abstract features. So we can conclude our experiment results that using the technique of Random Padding in the first two padding layers can always improve the performance of various deep learning models.

\begin{table}[]
\caption{Test errors (\%) with different data augmentation methods on CIFAR-10 based on VGG16 with traditional padding and VGG16 with Random Padding. \textbf{Baseline:} Baseline model, \textbf{RC:} Random Cropping, \textbf{RR:} Random Rotation, \textbf{RF:} Random Flipping, \textbf{RE:} Random Erasing.}
\centering
\small
\renewcommand\tabcolsep{9.0pt}


\begin{tabular}{|c|c|c|}
\hline
VGG16                     & Random Padding & Test error (\%) \\ \hline
\multirow{2}{*}{Baseline} & -              & 12.41 ± 0.08    \\ \cline{2-3} 
                          & \checkmark              & 10.54 ± 0.05    \\ \hline
\multirow{2}{*}{RC}       & -              & 10.54 ± 0.14    \\ \cline{2-3} 
                          & \checkmark              & 10.08 ± 0.09    \\ \hline
\multirow{2}{*}{RR}       & -              & 15.12 ± 0.05    \\ \cline{2-3} 
                          & \checkmark              & 9.82 ± 0.06     \\ \hline
\multirow{2}{*}{RF}       & -              & 10.37 ± 0.12    \\ \cline{2-3} 
                          & \checkmark             & 8.69 ± 0.07     \\ \hline
\multirow{2}{*}{RE}       & -              & 11.03 ± 0.08    \\ \cline{2-3} 
                          & \checkmark              & 8.89 ± 0.09     \\ \hline
\multirow{2}{*}{RF+RE}    & -              & 8.75 ± 0.13     \\ \cline{2-3} 
                          & \checkmark              & 7.75 ± 0.04     \\ \hline
\multirow{2}{*}{RC+RE}    & -              & 8.85 ± 0.07     \\ \cline{2-3} 
                          & \checkmark              & 8.34 ± 0.08     \\ \hline
\multirow{2}{*}{RC+RF}    & -              & 8.74 ± 0.06     \\ \cline{2-3} 
                          & \checkmark              & 8.26 ± 0.07     \\ \hline
\multirow{2}{*}{RC+RF+RE} & -              & 7.83 ± 0.09     \\ \cline{2-3} 
                          & \checkmark              & 7.21 ± 0.05     \\ \hline
\end{tabular}

\label{table3}
\end{table}

\subsection{Classification accuracy on different CNNs}
In this experiment, we use VGG16 as the benchmark model and apply Random Padding on the first two padding layers, and use CIFAR-10 as the test dataset. We chose four types of data augmentations, which are Random Rotation (RR), Random Cropping (RC), Random Horizontal Flip (RF) and Random Erasing (RE). The test error obtained by CIFAR-10 on the basic model of VGG16 is used as the baseline for this task. Then the effectiveness of our approach is evaluated by adding various data augmentations and combining with the Random Padding operation. 

As shown in Table \ref{table3}, the augmentation method of Random Rotation is not suitable for the dataset of CIFAR-10 on the VGG16, which means the use of Random Rotation makes the model's accuracy lower than the baseline. But after adding the Random Padding operation, the model's recognition rate on the CIFAR-10 test set exceeded the baseline, which indicates that the Random Padding operation can help the model learn features better. The model that combines a single data augmentation and the technique of Random Padding has a stronger learning ability than the model that only uses data augmentation. Therefore, the Random Padding operation is complementary to the data augmentation methods. Particularly, combining all these methods achieves a 7.21\% error rate, which has a 5.20\% improvement over the baseline.
\section{Conclusions}

In this paper, by investigating the learning of spatial information in convolutional neural networks (CNN), we propose a new padding approach named “Random Padding” for training CNN. The Random Padding operation reduces the extraction of features' positional information and makes the model better understand features' relationships in CNNs. Experiments conducted on CIFAR-10, CIFAR-100 and Fashion-MNIST with various data augmentation methods validate the effectiveness of our method for improving the performances of many CNN models. In future work, we will apply our approach to large-scale datasets and other CNN recognition tasks, such as object detection and face recognition.

\bibliographystyle{splncs04}
\bibliography{egbib}
\end{document}